# "Efficient Complexity": a Constrained Optimization Approach to the Evolution of Natural Intelligence

Serge Dolgikh[1][0000-0001-5929-8954]
Dept. of Information Technology,
National Aviation University, Kyiv
Lubomira Huzara Ave, 1

**Abstract.** A fundamental question in the conjunction of information theory, biophysics, bioinformatics and thermodynamics relates to the principles and processes that guide the development of natural intelligence in natural environments where information about external stimuli may not be available at prior. A novel approach in the description of the information processes of natural learning is proposed in the framework of constrained optimization, where the objective function represented by the information entropy of the internal states of the system with the states of the external environment is maximized under the natural constraints of memory, computing power, energy and other essential resources. The progress of natural intelligence can be interpreted in this framework as a strategy of approximation of the solutions of the optimization problem via a traversal over the extrema network of the objective function under the natural constraints that were examined and described. Non-trivial conclusions on the relationships between the complexity, variability and efficiency of the structure, or architecture of learning models made on the basis of the proposed formalism can explain the effectiveness of neural networks as collaborative groups of small intelligent units in biological and artificial intelligence.

## 1 Introduction

The emergence and development of natural intelligence happen at the conjunction of several thriving disciplines in modern science: information theory, biophysics, bioinformatics, computational, statistical and topological fields and aspects in mathematics and others. In this analysis, practical constraints imposed by the requirements of the physical existence of the system can be as important as general theoretical principles of the transfer of information and learning.

An essential observation in the theory of natural intelligence is that an intelligent system that is developing in a natural environment cannot be expected to have the information, description or a model, etc. of its environment that are accurate; detailed;

and complete at its prior state; therefore, this knowledge has to be acquired through some process of interaction with it.

This observation in the analysis of the processes and principles of the emergence and development of natural intelligence resonates with the studies of unsupervised artificial learning systems, a well-researched field in theoretical and applied computer science that explores the possibility and processes of acquisition of information from the interaction of a learning system with its environment that are not dependent on, and do not require significant prior knowledge about it.

In this work we attempted to outline a view on the progress of natural intelligence that combines information-theoretical and physical characteristics of the process of acquisition of information from the interactions with the sensory environment in the framework of the constrained optimization problem. This approach leads to the formulation of a general framework of the evolution of natural intelligent systems as a continuous process of fitting the information models of the sensory environments, guided by the optimization principle under the essential physical constraints as a basis for the construction of differentiated responses to the sensory stimuli the effectiveness of which is tested in empirical trials. Examination of the necessary conditions of a successful optimization strategy of traversing the extrema network of the objective function allowed to make some non-trivial observations on the physical structure or architecture of natural intelligent systems, specifically, the effectiveness of neural architectures in biological intelligent systems.

## 2  Prior Work

Information-based approaches in the theory of natural intelligent systems were introduced and developed since the groundbreaking works of Shannon and Shroedinger [1,2]. An extensive body of research has been compiled since in the active and expanding field that will be challenging to review with any fairness in a limited space, so we will focus on the directions and results that are directly related to the subject of our study: the theory of unsupervised learning, where essential information about the sensory environments can be inferred directly from the interactions with the environment (represented by samplings or "data"); and the thermodynamic and evolutionary approaches that attempt to formalize and describe the ability of natural learning systems to attain ever higher levels of attunement or fitness to their environments.

The ability of pioneering unsupervised generative models such as Restricted Boltzmann Machines, Deep Belief Networks [3,4] and related ones to create effective information models of simpler types of data is well-known and researched. Many effective approaches, types and architectures were examined since including neural models [5,6] that proved effective in modeling complex and realistic data such as high-quality images, music and others [7].

In the experimental studies, many important results were reported, such as the spontaneous concept learning experiment that observed the emergence of concept sensitivity on a single neuron level with an unsupervised generative neural model trained with massive sets of realistic images [8]. Distributions in informative representations of visual data were studied with generative neural models [9,10] pointing at the effect of geometric structuring in the generative representations. Studies [11,12] offered both broad and in-depth reviews of methods and approaches in unsupervised learning and dimensionality reduction.

Intriguingly, concurrently with these studies into artificial intelligent systems recent advances in the research in biological intelligence demonstrated the commonality of low-dimensional neural representations in the processing of sensory information by animals and humans [13,14]. These results suggest intriguing parallels between the learning processes of artificial and biological systems, perhaps guided by common laws and principles.

Following the foundational works of Shannon, many studies focused on the information processes in natural learning systems based on the framework of thermodynamics.

Studies [15-17] to name only a few, explored the principle of maximum entropy as the formalism of the description of the processes in the natural systems. Information entropy, introduced by Shannon can be seen as a natural measure of a "fitness" of a natural learning system to its physical environment and maximization of it, as a natural existential objective for such systems. However, there can be certain caveats in a direct application of this framework to the theory of natural intelligence; evidently, natural biological and learning systems exist in essentially open thermodynamic environments with strong flows of energy and materials, far from the equilibrium state.

As well, the essential existential constraints of natural systems cannot be ignored [18]. An important aspect of feasibility of any natural learning system, model, architecture, etc. is its ability to exist and operate within the essential physical constraints. In [19], the authors approached the problem of learning the essential information in the signal in the framework of constrained optimization. It can yield results on the optimal level of compression of the original signal and the generalization ability of the learning system where the correctness of the encoding is known either via a certain distance function (the rate distortion function) or another variable, such as a known category or class. We will attempt to show that the problem of natural learning can be formulated in the general case in terms of constrained optimization as well.

An active area of research lies in the plane of incremental variations and adaptations of the information model and structure (architecture) of learning systems: what processes guide their ability to model sensory environments of increasing complexity? Such processes require information models and architectures, for example, neural ones, to be flexible and adaptable that is, possess the ability to change incrementally [20,21], attaining progressively higher information fitness to their sensory environment while remaining with the limits of the essential existential constraints.

In view of the extensive body of research and the results accumulated over time in the field, open areas and questions remain. What are the connections between the general principles of learning in natural intelligent systems and the problems and directions in the theory and practice of learning, including data compression, dimensionality reduction and learning without prior knowledge? What principles and processes guide the ability of learning systems to evolve and adapt toward the ability to interpret progressively more complex sensory inputs and build differentiated responses that are both effective and efficient?

In this work, we attempted to approach these questions by first formulating the problem of natural learning as that of the constrained optimum: maximization of the objective function represented by the mutual information/entropy of the external states of the environment and the internal states of the learning system under the essential material constraints of natural learning systems that include, critically, memory; computation resources; energy; and other necessary physical resources. This formalism can yield non-trivial insights into the essential functions of natural learning systems including the ability to compress sensory stimuli as the basis for forming intelligent differentiated responses; and conceptualize them for efficient production of responses verified by empirical trials. It can also provide a conceptual basis and a formalism for the examination of the evolutionary ability of natural intelligent systems [22].

The rest of the paper is organized as follows: in Section 3, we discuss the statement of the constrained optimization problem of natural intelligence, including the formulation of the objective and essential constraints. Section 4 focuses on the formalization of the problem of generative learning and its connection to geometric categorization in informative representations of sensory data. Section 5 contains the formulation and analysis of the framework of evolutionary description of progressing natural intelligence as a traversal of the extrema network of the optimization problem under the objective of maximization of the information fitness and the results obtained in this plane of analysis. Section 7 contains the discussion of proposed approaches, results and their connections to other results and directions of research in the field.

## 3  Natural Intelligent Systems: Objectives and Constraints

As was commented earlier, natural intelligent systems are related to models in unsupervised learning field of computer science by the fact that they cannot rely on massive prior knowledge about their sensory environments.

One of the essential problems in the unsupervised learning relates to the ability of learning models to conceptualize sensory information, that is, factorize, group and compact it into a manageable set of common types, states or "concepts" that represent a cognitive model of the sensory environment. Correct interpretation of the general types or states of the environment of the learning system via sensory observations is an essential basis for constructing of effective differentiated responses to sensory stimuli i.e., those that maintain or advance the state of the system relative to its existential objective.

It can be concluded then that the problem of natural intelligence can be related, directly and closely, to interpretation of sensory observations and their samplings that will be generally referred to as "sensory data" as information models of characteristic types, concepts or states.

### 3.1 Essential Constraints of Natural Intelligent Systems

Natural intelligent systems are characterized by their ability to produce differentiated responses as a result of their interactions with the environment, while satisfying certain essential constraints governed by the physical reality of their existence. It can be argued that among those, an effective compression of sensory data is one of the requirements that are quintessential to the success of natural intelligent systems.

Indeed, physical resources available to them are limited and in the early stages of development, strongly constrained. Memory and computing power (compute) are the examples of the critical resources that are required and critical for construction of effective differentiated responses while strongly constrained by physical factors.

Memorizing previous sensory observations and their association with the internal states of the system can be critical for the construction of empirically effective responses: in the absence of such prior information, the correct interpretation and response to any sensory input would have to be relearned again and over. At the same time, realistic natural systems simply cannot "afford" to store raw sensory records due to the practical constraints of the available memory, energy and other essential resources.

Compute is another example of a strongly constrained resource: a realistic intelligent system cannot afford to build an entirely new response for each new sensory input within the practical constraints of resources and time. Then, it must attempt to group or "conceptualize" sensory stimuli into characteristic types, concepts or states; prioritize them i.e., rank by the perceived relevance to the existential objectives; and attempt to use already known general responses to similar stimuli.

These observations help to formalize the essential constraints of a developing natural intelligent system as:

1. The observed sensory samplings (data) $D$ must be compressed or compacted for storage in a representation or "embedding" $L$ of reduced dimensionality: $\dim(L) \ll \dim(D)$, while the information in $D$ that is essential for the production of correct differentiated responses is preserved to an acceptable extent.
2. Sensory data needs to be conceptualized, that is, effectively grouped into classes, types, concepts or states of similarity that can be associated with similar responses.
3. The constraints on the energy and other material resources must be satisfied at all times during the lifetime of the system.

It can be concluded then that for a natural intelligent system, compression of data that describes its sensory environment is a basic and critical practical necessity. Natural intelligence can emerge and develop only within the region of the internal parameters defining the system that satisfies the outlined constraints. Physically and practically feasible intelligent systems must, and have no choice but to compress sensory information as it relates directly to the essential constraints of their existence. Then, in its turn, an effective compression and conceptualization of sensory inputs can provide the basis for the construction of empirically successful differentiated responses to the sensory stimuli that maintain or advance the system toward its existential objective.

## 3.2 Objectives of Natural Intelligent Systems

As observed earlier, an intelligent system can be described by the ability to produce differentiated responses to the stimuli from the environment that advance its existential objective. As discussed previously, this ability requires some way of storing information about the earlier observations and trials.

The second important observation that was made is that this data (that is, samplings of the sensory environment) has to be compressed in such a way that it preserves the essential, for the intelligent system and its interactions with the environment, information about the sensory stimuli. If the compression results in a significant loss of information, it cannot be used to produce effective responses that will be successful in empirical trials.

Then, the objective function of a natural intelligent system can be defined as a measure of the coordination or correlation of the internal states of the system with those of its sensory environment, and the process of natural learning, as maximization of the objective function under the essential constraints of cognitive resources, energy and other necessary resources.

## 3.3 Natural Learning as a Constrained Optimization Problem

Let us consider sensory inputs in a certain space $D$ that can be expressed or "observed" by an intelligent system at its current state in certain observable factors $x$: the observable distribution $X \in D$, and its interpretation or representation by the system in a certain internal space $H$ described by "hidden" or latent factors $t$. The constraint of compression of sensory data with the preservation of information can be described by the class of encoding mappings $E = \{e\}\,(D \to L)$, the associated distribution $p_e(x,t), x \in D, t \in H$, and the mutual information [1] of the distributions $X$ and its representation, $E(X)$, $I_e(X, E(X))$.

Next, let the representation $H$ be a model of the sensory environment of some learning system $L$. Then, the objective of retaining the essential information about the sensory environment can be formulated as maximization of the mutual information $I_e$ of the states of the sensory environment in the inputs to the learning system and its internal states, over the space of encoding transformations $e \in E$, or, equivalently, distributions $p_e$ under the essential constraints of natural learning. The "correctness" of the encoding mapping (i.e., an association of the external states of the environment to the internal states of the system) and the distribution $p_e$ is verified by the empirical test, that is, the empirical success of the responses and actions produced by the system.

In the view of the discussion above, the optimization problem thus defined has to be constrained, specifically by:

— the critical constraints of the learning or cognitive resources such as the memory and computing power;
— energy and other material constraints such as physical materials required for training and operation of the system.

Then, the problem of natural intelligence can be defined as a case of the classical multifactor constrained optimization:

$$Max(I_e(e))\big|_{e:\Lambda}; \; d(e) \leq d_{max}; \; c(e) \geq c_{min}; \; \rho(e) \leq \rho_{max} \tag{1}$$

where $d, c$: the memory and conceptualization constraints, $\rho(e, g)$: the material resource constraints (physical resources, energy, etc.); $\Lambda$: the set of parameters that describe the encoding mappings, $E / p_e$.

The optimization problem in (1) then describes a case of inequality constrained optimization (Kunh-Tucker conditions, [23]). It is known that solutions of the problem (1) are the extrema, local or global of the Lagrangian functional:

$$L(I_e, \mu_{red}, \eta_{con}) = I_e(e) - \eta_{con}(d(e) - d_{max}) - \sum_k \mu_k(g_k(e) - c_k) \tag{2}$$

where $\eta_{con}$, $\mu_k$: the Lagrangian multipliers for the constraints of memory, conceptualization and other essential constraints in (1); $g_k(e)$, $c_k$: the constraints.

Next, one can make some practical observations on the application of the thus formulated problem statement to realistic natural intelligent systems.

Let us assume that the optimization variables ($e$) in (1) and (2) are described by certain parameters that define the full set of the learning parameters that describe the cognitive state of an intelligent system at a given point in its development, $\Lambda$.

With natural learning systems, the learning parameters are commonly divided in two classes. The physical constraints that are considered nearly immutable in the near-span learning process are described by the structure or "architecture" of the learning model: $A = \{ a_k \} \sim const$, whereas the training parameters $V: \{ v_j \}$ can be updated in learning to achieve optimums of the objective. The exact configuration of the model and the parameters is defined by the cognitive structure or architecture of the learning system for example, an artificial neural network, other artificial learning models or biological networks of neurons and synapses. Then, $\Lambda = \{ A, V \}$.

Next, a realistic natural system can be challenged to learn the exact value of the mutual information $I_e$ in (1) as it would require the explicit knowledge of the encoding distribution over all possible external and internal states $X, H$. Instead, in practical processes of learning, the unknown value of $I_e$ can be effectively approximated by a measure of accuracy or "fitness" $F_e(S)$ on a representative subset of samples $S$ in the original data space that can be calculated with a given configuration of the architecture and training parameters, $\lambda = (a, v)$ and a representative subset of samples in the input space, $S$. There are different approaches to this approximation as will be discussed in the sections that follow.

Then, with a fixed architecture and an effective approximation of the objective functional, the system can seek a solution of the optimization problem (2) via a Bayesian process [24] of updating the training parameters $V$ based on the difference (distance) between the prior (the prediction of the external states of the environment produced by the learning system from the inputs) and the posterior (the actual states discovered in empirical tests) achieving the minimization of the learning error and a solution, at least local, to the constrained optimization problem. In this interpretation, the problem of natural intelligence (2) translates into the standard problem of learning:

$$T_{opt}: Max\big(F_e(t)\big)\Big|_{S(D)} \qquad (3)$$

where $T_{opt}$: the optimal configuration of trainable parameters of the model that achieves the maximum of the measure of fitness or accuracy over a representative sampling of the observable distribution $D$, $S(D)$. Note that the introduction of an immutable architecture and the approximation of the objective function, the information entropy with the fitness measure effectively "hide" the constraints in the unconstrained problem (3) that can be seen as an approximation of (2).

## 4   Generative Learning and Conceptualization

In this section, we will discuss a well-established direction in self-supervised learning without prior knowledge that can be instrumental in the analysis of the constraints of compression and conceptualization. In this approach, the ability of the system to learn and conceptualize sensory inputs stems from the capacity to generate the observable distribution from its compressed, "encoded" form. Models of generative self-supervised learning are thus trained to reproduce the input distribution with high accuracy and precision by imposing the incentive to reduce the error or distance between samplings (i.e., sets of sensory inputs) and generations produced by the model in the process of learning.

### 4.1   No "Natural Limit" in Unsupervised Data Compression

Due to critical resource constraints, dimensionality reduction is essential or even critical for empirical success of natural learning systems. It has been studied at great depth and length with a large number of insightful results obtained to date including those mentioned in the review section. An essential observation that will be made from the outset of this analysis is that in the case of unsupervised dimensionality reduction, that cannot rely on prior knowledge of the essential characteristics of the sensory data, there is no theoretical, information theoretical or other general cause or principle that can determine or point to the best or optimal level to which the data needs to be compressed.

The argument first observes that any generative data compression of an original sampling, distribution or data $D$, that is capable of restoring it from a compressed representation $R$ via a generative transformation $G: R \rightarrow D$ would result in a loss of information about if the dimensionality of the representation is lower than that of the original distribution, excluding special cases as degenerate parameters, limited region of the distribution, etc.

This conclusion follows from the results on the invariance of topological dimension (Brouwer's invariance of domain and the related). Indeed, if the perfect compressed representation existed in the lower dimension, there would exist a homeomorphism between the topological distributions of the data in the original space and its compressed image in the latent space, of two different topological dimensions. That supposition would then contradict the results on the invariance of the topological dimension.

Then, a trivial case of the perfect, zero-loss representation is always available if the dimensionality restriction is lifted with the identity transformation $E(D \to D): E(x) = x$, where *D*: the original data (a distribution of data points that represent sensory stimuli in the space of the original observable parameters).

Thus, between the identity transformation and a compressed representation *R* in a latent space *L*, $R(D \to L)$, $\dim(L) < \dim(D)$, any non-trivial reduction of dimensionality would result in a loss of some information about *D*, defined as the non-existence of a mapping that allows to restore *D* from *L* precisely:

$$G(L) \simeq D \Rightarrow \text{Dim}(L) \geq D \tag{4}$$

where *G*: the generative transformation, and excluding aforementioned exceptions.

From this result, as the optimal level, dimension or criteria of the compression of data cannot be formulated in general terms based only on information theoretical principles, there follows an essential for natural learning systems observation that *the constraints and incentives of dimensionality reduction in their information models have to be dictated by the practical conditions of their existence and interaction with the environment*.

These conditions, as discussed earlier, can be expressed formally as the essential constraints of natural learning and they determine the balance or trade-off between the preservation of the essential information in the reduced representation and the level or compression.

### 4.2 Constrained Optimization in Generative Learning

As in Section 3.3 we consider sensory input space *D* and its compressed representation in the latent space *L* described by latent factors *t*, encoding transformations $E(D \to L)$, the associated distribution $p_e(x, t), x \in D, t \in L$, and the mutual information of the distributions *X* and its latent image, $E(X)$, $I_e(X, E(X))$.

In the generative approach, the intelligent system must also have the means to restore the compressed representations to the input space to compare the stored information with new inputs, and to verify the correctness of the encoding function. This function can be described by the class of generative transformations: $G(L \to D)$, the distribution $p_g(t, x), x \in D, t \in L$, and the mutual information $I_g(t, G(t))$.

Then, following the approach outlined earlier in Section 3.3 and the concept of generative learning outlined above, the fitness objective of generative learning can be defined as maximization the mutual information of the sensory data in the input to the learning system and distribution generated distribution: $I_r = I(X, X'=G(t))$ over the space of encoding and generative transformations described by the tuple (*e*, *g*), $e \in E$, $g \in G$) again, under the essential constraints of natural learning:

$$Max\bigl(I_r(e,g)\bigr)\Big|_{e,g:\Lambda};\ d(e,g) \leq d_{max};\ c(e,g) \geq c_{min};\ \rho(e,g) \leq \rho_{max}$$

As in (1), *d*, *c*: the memory and conceptualization constraints, $\rho(e, g)$: the combined resource constraint (physical resources, energy, etc.); *Λ*: the set of parameters that describe the encoding/generative tuple (*e*, *g*) with the Lagrangian of generative learning:

$$L(I_r, \mu_{red}, \eta_{con}) = I_r(e,g) - \sum_k \mu_k(g_k(e,g) - c_k) \tag{5}$$

where $\mu_k$: the Lagrangian multipliers for the constraints, $g_k$(e, g), $c_k$: the constraints of dimensionality reduction, conceptualization and other material constraints.

Similarly to what has been outlined in Section 3.3 the optimization variables (*e*, *g*) of generative learning in (5) are defined by the architectural/structural and training parameters: $\Lambda = \{A, V\}$.

An effective approximation of the mutual information in (5) can be modeled by a measure of generative accuracy $F_g(S, G(E(S)))$ on a representative subset of samples *S* in the original data space, i.e., the distance between the original sampling and the result generated by the model in the metric of the input data space ("the generation trick"); it can be calculated readily with a given configuration of the architecture and training parameters, $\lambda = (a, v)$ and a representative subset of samples in the input space, *S*.

Then, as in (3), Section 3.3 a solution of the optimization problem (5) can be sought via a Bayesian process of updating the training parameters *V* based on the difference (distance) of the prior (the original set of samples) and the posterior (the generation produced by the learning model):

$$T_{opt}: Max\left(F_g(t)\right)\Big|_{S(D)}$$

where $T_{opt}$: the optimal configuration of trainable parameters of the model that achieves the maximum of the measure of generative accuracy over the distribution of the original data; *S(D)*.

The statement above describes the standard formulation of the problem of self-supervised learning [11,12]. Note again how the introduction of the immutable architecture and the generation trick effectively "hide" the essential constraints in the unconstrained problem of unsupervised learning (5).

### 4.3   Generative Data Compression and Geometric Conceptualization

In this section we will attempt to provide arguments on the constraint of conceptualization, that is related to the ability of the system to factorize or classify stimuli into common types, specifically, that in some cases in some cases it can be effectively realized by that of a strong dimensionality reduction. We will consider the following lemma of geometric conceptualization:

Under the conditions of:

1. Learning success: the learning system is capable of achieving generative accuracy above a certain minimum: $A(D, G(D)) > A_{min} = \alpha$
2. Generalization: the accuracy is preserved across all and any representative set of samples in the original data space: $\forall S(D): A(S, G(S)) > \alpha$.
3. A strong dimensionality reduction to a low dimension $d_L$, that allows simplified topological classification of the data distribution manifolds.
4. Conceptualization or factorization of data in the observable space: the original distribution is comprised of characteristic types of similarity: $D = \{C_1, .. C_k\}$, $C_k$: the classes or types of similarity in the observable space. Note that neither the types nor their observable or latent distributions are known to the learning system at prior.
5. Immutable structure (architecture) of the learning model in the stages of training and trials.

generative models with well-separated (as defined below) regions of the latent distributions of the types of similarity (concepts) are prevalent in the ensemble of learning models of the same architecture trained on the same representative input sample.

**Outline of proof**

We will consider the simplest case of the composition of data with minimal number of types: two, $D = \{A, B\}$ and a generative intelligent model $M$ that satisfies the conditions of the lemma.

Next, we consider a representative sample $S \subseteq D$ and the latent region $X$ of the strong intersection (intermixing) of the latent distributions of the types $A$, $B$ in the latent image of $S$. The condition 3 (dimensionality reduction) ensures that the latent distributions $L_A = E(A)$, $L_B = L(B)$ are well defined topologically (for example, as one or finite number of compact manifolds of dimension) and therefore, so is their intersection. The condition of strong intersection means that it contains significant populations of both classes, in a general manner, that is, for all representative sets of $D$. Formally, $\forall\, g \subseteq X\ P_A(g) \sim \beta\ P_B(g) > 0, P_A, P_B = Card(L_A \cap g), Card(L_B \cap g), \beta: const$. Any subregion of $L_A \cap L_B$ that does not satisfy the condition of strong intersection is removed from $X$.

For the simplicity of the abridged version of the proof presented here, it will be assumed that the populations of the types in the observable data, and in the condition of significant intersection region $X$ are approximately equal: $\beta \sim 1$. The proof can be extended to the general case straightforwardly and will be provided in another study.

Now, assuming $X \neq \emptyset$ let us take a latent point $y \in X$ with the image in the observable space $G(y)$ is of type $A$: $G(y) \in A$, where $G$: generative transformation of $M$. Then, applying the condition of strong intersection to an arbitrarily small neighborhood of $y$, $\varepsilon(y)$ one can conclude that it has to contain approximately equal populations of latent positions of classes $A$, $B$.

It is known that models of finite complexity, including deep neural networks [25] commonly have finite resolution, described by the Lipschitz constant, $l$: $\|G(y), G(x)\| \leq l\ \|y, x\|$. Then, for sufficiently small $\varepsilon$ it follows that $G(x) \in A\ \forall\, x \in \varepsilon(y)$. Then, based on the conclusion of equal populations of classes above, it would follow that the model produces an erroneous generation in approximately ½ of the population of $\varepsilon(y)$. Finally, as $y$ represents an arbitrary position in $X$, this conclusion can be extended to the entire region. Where $w_g$ is the generative error of the model,

$$w_g(S) \geq w_g(X) \sim \frac{1}{2}\ P(X) = \frac{c}{2}\ \frac{m(X)}{m(E(S))}$$

where $P$: the population of the latent region $X$ with an observable set $S$, $m$: a measure of volume in the latent space; $c$: a constant.

Then, for the overall generative error of the model on $S$ one obtains:

$$w_g(S) \geq \frac{c}{2} \mu_X$$

where $\mu_X$: the relative volume of $X$ in the latent space, $\frac{m(X)}{m(E(S))}$.

Now, the conditions 1 (accuracy) and 2 (generalization) of the lemma require that $w_g(S) \sim$ const and $w_g(S) \leq 1 - A_{min}$. Then,

$$\mu_X \leq \beta \, (1 - A_{min}), \; \beta: \text{const} \tag{6}$$

connecting the generalized accuracy with the relative measure of the latent region of the strong intermixing of the general types in the observable sample.

This argument can be extended straightforwardly to the arbitrary number of general types. It follows then that under the conditions of the lemma, successful generative models with latent representations of sufficiently low dimension must prefer conceptualized latent distributions of the general types (concepts), with minimal intersection of their latent regions of distribution.

One can note that all of the conditions were essential in the proof. Another condition that was assumed implicitly is that the essential characteristics of the original distribution $D$ do not change during the training or any of the trial stages.

With practical implementations of intelligent systems, the assumption of infinitesimal continuity of the latent space on which the outlined proof was based can be substantiated by an observation based on the condition of generalization of the lemma. Even if, with a finite representative set $S$, an arbitrarily small neighborhood may not be expected to contain the latent positions of physical samples in $S$, it can still contain those of samples in other such sets, $S_1, \ldots S_k$. Then, the argument in the proof that all such samples in a sufficiently small neighborhood would be classified to the same type still holds.

### 4.4 Empirical Results in Generative Geometric Conceptualization

The results on geometric conceptualization of generative representations are supported by a number of published experimental results obtained with different types of data and generative architectures. The results [9,10] demonstrated and described "disentangled" representations of visual data with several datasets of images. The emergence of concept-correlated structures in unsupervised generative learning with very large sets of images was reported in [8]. Geometric and topological characteristics of generative representations were studied in [26] with several datasets of images describing a well-defined continuous structure of concept regions.

These results provide direct experimental support for the conclusions on geometric conceptualization of generative representations outlined in the preceding sections. Recent results in experimental neuroscience point at the ubiquitous character of low-dimensional representations of sensory stimuli in biological organisms, including visual, olfactory and audio signals in animals and humans [13,14]. It can be conjectured that this observed effect can be related to the higher effectiveness of low-dimensional generative representations in identifying characteristics types, or concepts in the sensory data.

### 4.5 Incidental Advantages of Deep Data Compression

It may be worth noting certain additional "incidental" benefits of deep compression of sensory data for natural learning systems that can be important or even critical for the success of conceptualization:

1. Compressed distributions of lower dimensionality can have simpler topological structure. A number of known results in topology point to this observation: Smith conjecture breaks in $d > 3$; Milner conjecture breaks in $d > 5$; classification of compact manifolds is solved in $d < 4$; algorithmic PL homeomorphism problem for compact manifolds solved in $d < 4$; and other results ($d$: the dimension of topological space). All of these results can have a direct connection to topological characteristics of distributions in the low-dimensional representations of sensory data and the ability of a learning system to determine the concept structure in them.
2. Significantly reduced dimensionality of internal representations can be essential for some computational methods in the task of conceptualization such as clustering, that can require progressively more time and computing power with higher dimensionality of the data or even become intractable [27].
3. If strong reduction of dimensionality is successful (i.e. avoids significant loss of essential information in the original distribution) it can point at the original data being strongly redundant. There can be downsides in attempting conceptualization directly with strongly redundant data such as a possibility of overrepresentation of some factors.

All of these factors can be linked directly and significantly to the effectiveness of conceptualization of sensory observations: as already commented, a critical need and constraint for any realistic natural intelligent system.

### 4.6 Practical Corollaries of the Geometric Conceptualization

For an illustration of the importance of the dimensionality constraint and geometric categorization for practical intelligent systems let us consider an example of two intelligent systems. One (A) was trained to achieve lower accuracy (and therefore, higher information loss) for example, 70% but more effective conceptualization, possibly due to producing an effective low-dimensional representation: 80%; the other system (B) achieves near-perfect accuracy, but less effective conceptualization: 95% and 70%, respectively.

Then, assuming that correct responses are associated with the identified general types of sensory inputs with perfect accuracy, A would produce 80% of correct responses, whereas the second one (B), 70%. Then model A will be selected in empirical trials due to the higher effectiveness of its responses.

This simple example underlines the observation that the objective for practical intelligent systems is not an unconditional and unconstrained maximization of the correlation between the sensory information and its internal representation, commonly described by the standard problem of unsupervised learning; but rather, an effective conceptualization of it with preservation of essential information and within the subspace of the parameters of the system that satisfies the essential constraints (2). Strong dimensionality reduction thus plays a key role in satisfying two of the essential constraints of natural intelligence: that of limits on the critical resources, the memory and the compute; and effective conceptualization. The result on geometric categorization is important in this perspective as it allows to effectively replace the conceptualization constraint with that of the dimensionality of the internal model of the sensory data that can be defined and described explicitly in the architectural parameters of learning models.

An essential corollary of the result on the geometric conceptualization of generative representations is that the conceptualization constraint that can be challenging to express explicitly in the architectural parameters of the system can be effectively replaced, in some cases at least, with the constraint on the effective dimensionality of the internal representation space. Then, both of the constraints of the memory and conceptualization in (2) and (5) can be described by a strong constraint on the dimensionality of the internal representation:

$$d(\lambda) \leq d_{max}; \ c(\lambda) \geq c_{min} => d(\lambda) \leq d_{con},$$

$d_{con}$: the constraint of effective compression and conceptualization.

This observation allows to obtain the explicit form of the Lagrangian (3) based on the results on inequality-constrained problems:

$$L(F, \mu, \eta) = F_e(\lambda) - \mu \left( d(\lambda) \leq d_{con} \right) - \eta \left( \rho(\lambda) \leq \rho_{max} \right) \tag{7}$$

where the objective function $F_e(\lambda)$ and the constraint operate in the space of architectures $\Lambda$, with the Kuhn – Tucker conditions on the solutions (and similarly, for the case of generative learning, (5)).

It can be noted in conclusion that generative learning and conceptualization considered in this section present one possible solution and strategy for a learning system to comply with the essential constraints of natural learning, including the compression of sensory inputs and conceptualization. Yet, there are no reasons to expect it to be either the general one or the only possible. Other possibilities, for example, effective one-way dimensionality reduction methods and strategies whose effectiveness, including accuracy and precision, can be established empirically in trials. From here onwards it will be assumed that the learning system is able to achieve the compression and conceptualization objective under the constraints via some strategy and the approach outlined in (1)-(3), (7) will be used from this point on.

## 5 Evolving Natural Intelligence

### 5.1 Information Fitness and The Information Model

Let us return to the optimization problem (2) with the objective function $I_e$, the mutual information of the observable (sensory) and the encoding distributions.

An intelligent system that has attained a local objective maximum under the constraints as described earlier can associate input stimuli $x$ in the observable space to their encoded positions $t$ in the space of internal factors that describe the state of the system satisfying the constraints of the problem (2). Then, with sufficient empirical trials, the following probability distributions can be estimated empirically:

— $P_s(x)$: the empirical probability distribution of the sensory inputs $x$,
— $P_i(t)$: the empirical probability distribution of the internal variables, $t = E(x)$,
— $P_m(t, x)$: the empirical joint probability distribution of the sensory inputs and their internal images or representations.

With these empirical variables, the objective function of the optimization problem, the mutual information between the internal variables of the system and the sensory stimuli $x \in X$ can be calculated as [1]:

$$F = \sum_x \sum_t P_m(t,x) \log \frac{P_m(t,x)}{p_s(x)\, p_i(t)} \tag{8}$$

The objective function above can be interpreted as the "information fitness" of an intelligent system. It can be seen immediately that it is directly related to the empirical success of the responses produced by the system.

Indeed, assuming for now that the system can construct perfectly effective responses based on the identified internal state of the stimuli (as the model of production of the responses will be discussed elsewhere), the empirical effectiveness of the response will be determined by the correctness of the association between the sensory stimuli and the internal states of the system that is used to construct the response, and this is exactly what the information fitness characteristic is a measure of $F$.

The joint probability distribution $P_m(t, x)$ in (8) will be referred to as the information model of the intelligent system:

$$M = P_m(t,x) = P_{[T,X]} \tag{9}$$

As noted above, it describes the empirical correctness of the interpretation of the observables $x$ by the internal variables of the system $t$, verified empirically by the effectiveness of the responses produced by the system. An discussed earlier, the information model can be expressed by a parametric function of the observable factors, $\mu(\Lambda, X)$:

$$M: \mu_{v,a}(X) \to E(X)$$

Next, based on the result of the lemma of geometric conceptualization in Section 4.2, we will assume that the distributions $X$, $E(X)$ in the observable and latent spaces are conceptualized: $X = \bigcup_C C_k$, $E(X) = \bigcup_j K_j$, $C_k$: general types of similarity of sensory inputs; $K_j$: internal concepts or "states" of the system associated with distinct regions in the latent space, according to the lemma of conceptualization. Then, as easy to see, the definitions (8) and (9) can be rewritten in terms of characteristic classes $C$ and internal states $K$ that will be referred to as the external states of the environment and the internal states of the system, respectively. To eschew cluttering of symbols, the variables $x$, $t$ will denote both external factors and states, and internal (latent) factors and states, respectively. Where the distinction is essential and not obvious from the context an explicit note will be made.

### 5.2 A Two-State Model Intelligent System

For an illustration of the definitions given above, let us consider information models $M_1$ and $M_2$ of a simplest type that interpret a single observable $v$ mapping it to a single internal variable, "viability" of the system $t$ with two possible states. Such a model can be realized with a single intelligent unit, such as a diode or a neuron.

$$if\ x \in r_a: t = True\ (\text{"friendly"});\ else\ t = False\ (\text{"hostile"})$$

$r_a$ being the viable or hospitable range of the observable $x$.

We consider a model $M_1$ of this type that maps the observable to the internal state mostly correctly, producing correct interpretation with the probabilities 0.9/0.1 in each of the ranges of the observable $x$; whereas the other model, $M_2$ fails to learn or has yet to, producing near-random responses. The information models $M_1$ and $M_2$ are described by the probability matrices $(t, x)$, Fig.1.

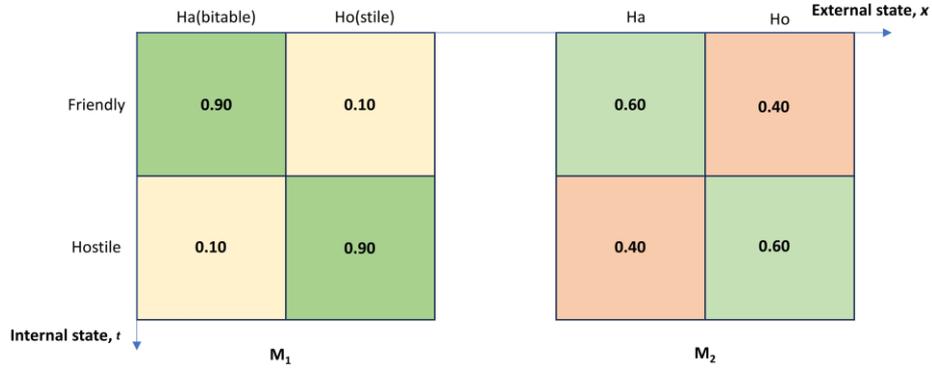

**Fig. 1.** A two-state intelligent system and information model.

Then, taking as an example the empirical distribution of the observables $p_s = (0.7, 0.3)$ one can calculate the values of the fitness function for the models as: $F(M_1) = 2.28$; $F(M_2) = 1.26$. A model with the best fitness possible in this case, represented by a diagonal probability matrix has, under these settings, the maximum possible fitness value, $F(M_p) = 2.99$. Indeed, it can be seen that the models that produce responses in a better correlation of the internal and external states have higher value of the information fitness.

The correctness of the mapping of the external states of the environment to the internal ones by the intelligent system is verified in empirical trials. Indeed, in the example above, assume that the systems can produce the simplest differentiated response $R$: "if the internal state is "friendly": remain in place; else, move in a randomly chosen direction for a random distance; repeat".

Then, in an environment that has a sufficient prevalence of habitable regions, the simple intelligent system described above, with a correct or "fit" information model will be able to survive and even thrive. On the other hand, an essentially wrong, "broken" mapping of the external and internal states can be detrimental to its survival.

The characteristics of information fitness $F$ and information model $M$ of an intelligent system thus define its ability to produce effective responses to sensory stimuli in a given empirical environment incorporating all essential constraints discussed earlier.

### 5.3 Evolving Natural Intelligence

In the framework of the constrained optimization and its connections to the architecture and training of learning models in natural environments outlined in this work, one question remains: how, through which processes can natural intelligent systems acquire the ability to process, interpret and utilize sensory data of increasing complexity for the construction of more complex and effective responses, solving the optimization problem in (2) in progressively more complex sensory environments?

It can be noted that an emerging natural intelligent system may not, and commonly, would not have detailed information about its sensory environment and for that reason, would have to acquire it through some process of interaction with it. It may not be able to implement complex algorithms to seek the solution to the optimization problem for the same reason, or due to the essential constraints.

The two planes where nature is not constrained is the time and the number of trials. Rather than employing complex and often theoretically intractable approaches to multidimensional non-linear constrained optimization problem, nature can choose the evolutionary path of incremental variations and adaptations with selection according to certain essential criteria.

As an illustration, let us take the example of perhaps one of the simplest intelligent system possible considered in the preceding section. It is capable of mapping two or more intervals of a single sensory observable to a binary internal state. As was shown, even a simple system of this kind is capable of surviving and even thriving in certain simple sensory environments.

By adding one more intelligent unit, for a two-neuron information model, an intelligent system would acquire the ability to model significantly more complex single channel input distributions including linear functions; and/or approximate two-channel ones, including logical functions and others [28]. Such a system would be capable of alternating or selecting different modes of responses depending on the state of external observables.

An incremental adaptation of a different type: convolutional vectorization [29], a function or component that can translate visual signals to scale-invariant numerical vectors and vice versa can be considered a processing improvement that preserves higher information content of the sensory inputs; by combining it with a small number of intelligent units (from as low as 2-3, [26]), an intelligent system can acquire the ability to recognize some simple geometric shapes such as circles, triangles, and others . Clearly, an ability to conceptualize and differentiate even simple visual forms can be essential for survival of simple intelligent species.

The sequence of incremental adaptations in the architecture of artificial neural models was shown to produce a noticeable improvement in the ability to recognize characteristic types (concepts) in the visual data, up to complex visual forms such as handwritten digits [22]. The sequence of incremental architectural adaptations capable of conceptualizing sensory data from the simplest, binary variable mapping to complex visual data comprising up to $10^2$ distinct conceptual states is illustrated in Fig.2.

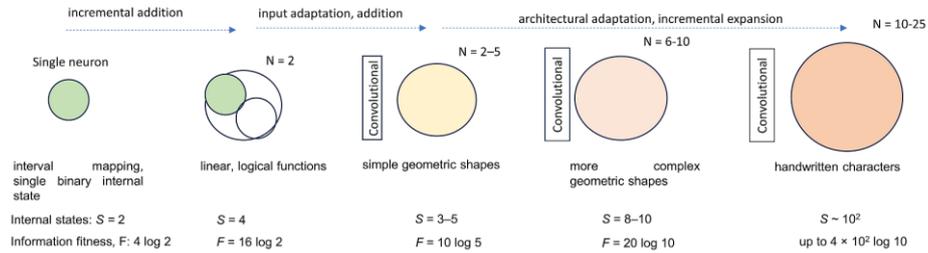

**Fig. 2.** Chain of incremental architectural adaptations characterized by improvement in information fitness; *N*: the number of intelligent units.

On the assumption of the perfect mapping of the observable states to the internal ones, a chain of adaptations would have produced a significant improvement in the information fitness of the information models in Fig.2: from $F \sim 2.99$, the two-state model system up to $F \sim 9.2 \cdot 10^2$, some simple visual sensory environments, as a result of the sequence of cognitive architectural adaptations in this example.

On the basis of the comments made earlier on the initial state and the constraints of a developing intelligent system and the example of architectural adaptations above, it can be conjectured that a natural intelligent system can seek solutions to the optimization problem (2) via an evolutionary process of incremental adaptations in the space of parameters that define its structure (architecture) with selection by the value of the objective function, that is, the information fitness of the internal and external states verified empirically and within the essential constraints of the problem.

### 5.4 Evolutionary Formalism of Natural Intelligence

An evolution in a space of states of a system can be defined by incremental variations with selection by the fitness to the objective progressing toward an optimum under the defined criterium. Thus, it can be formally described by a quad tuple of components $E_v = (A, V, R, K)$ [30] where:

— *A*: the space of possible states of the system which are formally describable, with explicit descriptions represented by certain variables { *a* };

- *V(a)*: a subspace of incremental variations of the states that are allowed within the constraints of the system;
- *R(a)*: a certain factor of selection that can obtained or calculated from the current state of the system and its interactions with the environment;
- *K(a)*: the selection criteria (and the process) on the basis of the selection factor.

An evolving population $P(t_0) = \{ p_k \}, p_k \in A$ at the initial point $t_0$ of the evolution can be described by a certain distribution of possible states of the system satisfying the constraints. Then, at the next evolutionary point $t_1$, the evolved population can be described as: $P(t_1) = \{ p_k + v_k\}, v_k \in V(p_k)$ with the range of the selection factor $R(t_1) = R(p_k + v_k)$.

$$R(t_1) = R(p_k + v_k) = R(p_k) + v_k \nabla R^k, \nabla R^k = \left.\frac{\partial R}{\partial v_k}\right|_{p_k}$$

In evolutionary models observed and studied, it is presumed the selection process *K* selects the individuals in the population with "better" characteristics of the selection factor, *R*. In the examples of natural systems considered earlier, individuals with more effective empirical responses can survive longer, expand to a larger area and so on.

In application to the problem of natural intelligence, the evolution can be described by the following components: $A = \Lambda$, the architectural parameters of the information model; $V = \Delta(\lambda)$, the subspace of possible incremental variations of the architecture under the constraints of natural learning; the information fitness *F* of the information system defined earlier; and the empirical selection mechanism, *K* that selects the individuals in the population that achieve higher information fitness. Then the evolution process of natural intelligence can be defined as:

$$E_{nat} = (\Lambda,\ \Delta,\ F, K) \qquad (10)$$

Next, one can describe the necessary conditions of a successful evolution as:

1. The solutions of the optimization problem, global and local, exist.
2. At least in some points or regions in the system architecture space the subspace of incremental variations is not empty: $\exists \lambda: \Delta(\lambda) \neq \emptyset$.
3. Transitions between the local extrema $\lambda, \lambda'$ of the objective function (2) are possible for some regions in the system architecture space: $\exists \lambda, \lambda': p_{tr}(\lambda, \lambda') > 0$.

where $p_{tr}(\lambda, \lambda')$: the probability of transition between the architecture states $\lambda, \lambda'$ that satisfy the constraints of the problem.

Under these conditions, the evolutionary process just described can effectively traverse the system parameter space finding solutions with the improvement in the fitness factor. The sequence of architectural adaptations discussed earlier in Section 5.2 demonstrated an example of such an evolutionary path or trajectory in the neural architecture space.

### 5.5 Evolutionary Dynamics of Natural Intelligence

In the framework defined above let us consider the probability distribution $p(F)$ of a certain information model described by architectural parameters $\lambda$ by the information fitness it achieves (the F-distribution). It can be an actual distribution in the population of intelligent systems of similar architecture, $p(F) = \frac{n(F)}{N}$, $n(F)$ being the sub-population with the fitness of $F$ in the total population $N$, or the probability of an individual model to attain certain range of fitness, $p(F) = \frac{\partial P(F)}{\partial F}$ following training. Then, in the framework of the variational evolution defined by (10), the change in the fitness distribution as a result of an architectural variation $\delta\lambda$ at the next evolutionary point $t_{k+1}$ can be written as:

$$\delta p(F, \delta\lambda, t_{k+1}) = \Xi(F, \lambda, \delta\lambda)$$

where $\Xi(F, \lambda, \delta\lambda)$ is the adaptation functional that defines the relation between the architecture and the fitness factor; it is therefore, specific to the problem.

Next, let us consider the subspace of all possible incremental structural variations at a given architecture point $\lambda$: $\Delta_f(\lambda)$. Then, the sum of the F-distributions of the possible architectural variations at $\lambda$ is:

$$\delta p(F, \lambda, t_{k+1}) = \sum_{\Delta_f} \Xi(F, \lambda, \delta\lambda) \, q(\lambda, \delta\lambda), \qquad (11)$$

$q(\lambda, \delta\lambda)$: the probability of the incremental variation $\delta\lambda$ at $\lambda$.

The architecture described by $\lambda$ will be defined as "adaptable", if it can improve the fitness distribution via a structural variation: $\exists F: p(F) = 0; \ \delta p(F, \lambda, t_{k+1}) > 0$. Then, $\exists \delta\lambda: \delta p(F, \delta\lambda, t_{k+1}) > 0$; the variation $\delta\lambda$ is then defined as an architectural adaptation with an improvement in the selection factor.

The role of the selection function $K$ in (10) is to select or "prune" the evolved distribution (11) according to some existential objective or the imperative. In the framework described here it can be the improvement in the information fitness of the evolved distribution (or population) that can be defined in different ways in a specific problem. Then, only the variations that satisfy the criteria imposed by $K$ in the subset of variations $\Delta_a(\lambda)$ are selected in the resulting distribution. As a result,

$$\delta p(F, \lambda, t_{k+1}) = \sum_{\Delta_f} \Xi(F, \lambda, \delta\lambda) \, q(\lambda, \delta\lambda) \, K(\delta\lambda) = \sum_{\Delta_a} \Xi(F, \lambda, \delta\lambda) \, q(\lambda, \delta\lambda) \qquad (12)$$

The last equation of the evolutionary dynamics presumes that the selection process eventually propagates the evolved distribution to the general one:

$$p(F, \lambda', t_{k+1}) = \delta p(F, \lambda, t_{k+1}); \ \Delta_a \neq \emptyset \qquad (13)$$

where $\lambda' \in \Delta_a$: the new base or standard architecture that was selected in the process of architectural adaptation. As a result of transition (13), the distribution $p(F)$ shifted toward higher values of the information fitness by locating a new local maximum of the optimization problem (2) in the architecture space, $\lambda'$ while within the constraints of the problem.

An equivalent formulation of the dynamical equations (11) – (13) in terms of the populations of individuals can be given straightforwardly.

## 6 Applications of Constrained Optimization Framework

### 6.1 "Efficient Complexity": Variability-Cost Balance in Evolving Natural Intelligence

As stated earlier in Section 5.1 the necessary condition of successful transition to a higher fitness can be written as:

$$\exists \lambda: \Delta_a(\lambda) \neq \emptyset, |\Delta_a(\lambda)| > 0 \tag{14}$$

that implies: $|\Delta_f(\lambda)| > 0$, as $\Delta_a(\lambda) \subseteq \Delta_f(\lambda)$

Whereas a closer analysis of the explicit form of $\Xi(\delta\lambda)$ in application to intelligent systems will be considered in a dedicated study, it can be noted that the success of evolution described in (11)–(13) is strongly determined by the volume or "richness" of the subspace of incremental variations, $\Delta_f$. Indeed, a small, "narrow" variational space may suppress the variance in the distribution $\delta p(F, \lambda)$ reducing the potential fitness gain from architectural adaptations.

As was noted previously, in the context of the optimization problem (2), architectural adaptations of an evolving natural intelligent system correspond to the local extrema of the Lagrangian of the information fitness optimization problem (5). Then, the essential condition of the evolution (10), 1–3 for a natural system can be interpreted as non-trivial constraints imposed on the architecture by the ability to migrate between the local extrema of the objective function.

Indeed, assuming that there exists a characteristic scale, $L_{ex}$ that represents the characteristic minimal distance in the architecture space between the adjacent extrema of the objective function (2), this factor essentially, dictated by the characteristics of the environment and can be thought of as a characteristic scale of the extrema network of the problem. Let $H_{ex}$ be the cumulative energy cost (including all material constraints) of the transition from the current architecture $\lambda$ to the adjacent local maximum of the objective function, $\lambda'$. Then, the necessary conditions of evolution (10), 1–3 require, simultaneously, the volume of the local subspace of incremental variations $\Delta_f(\lambda)$ at $\lambda$ to be sufficiently "large" to contain the new extremum position of the objective function; while at the same time, be "efficient" with respect to the energy required for the incremental change of the architecture, satisfying both the constraints of the problem (the weaker constraint) and the possibility of the transition to the new state (the stronger constraint), $p_{tr}(H_{ex}) > p_{min}$.

While an explicit formulation of this condition requires a detailed analysis of the transition probability $p_{tr}$ and will be attempted in another work, it can be interpreted generally as "*efficient complexity*" of the structure (architecture): maximizing the variability of the architecture while minimizing the energy cost of architectural variations, or maximizing the energy gradient of architectural adaptations:

$$G_H(\lambda) = \frac{\partial \lambda}{\partial H} \to max \qquad (15)$$

A way to realize such a solution in naturally feasible systems would be to construct them from small intelligent units with a minimal energy/resource cost so that incremental architectural variations can result in non-trivial improvements in learning (as illustrated earlier in Section 5.2). Then, the ubiquity of neural architectures in natural intelligence may be due to more than it being an effective type of intelligent model/strategy but in fact, a natural solution to the constraints of evolvability of cognitive architectures. The condition (15) then assures the ability of the learning system to traverse ("hop over") the extrema network of the optimization problem along a trajectory of solutions leading to ever higher information fitness with the environment.

### 6.2   Incremental Addition of a Cognitive Structural Unit

As another illustration of applicability of the framework of constrained optimization, we will consider the scenario of incremental addition of a cognitive structural unit into an intelligent system, such as a neuron or similar. For simplicity, we will focus on one kind of essential cognitive resource, memory, or conceptualization. We will also analyze here the case of strongly constrained resource, the boundary of the constraining condition. The general case will be analyzed in more detail in another study.

Let the parameter $\lambda$ describe the structural composition (cognitive architecture) of an intelligent system, with an incremental variation, described by an addition of a structural unit, $\delta \lambda \ll \lambda$. The Lagrangean conditions of an extremum of the constrained optimization problem can be written, from (2) as:

$$\frac{\partial I_e(\lambda)}{\partial \lambda} - \eta_c \frac{\partial d}{\partial \lambda} - \mu \frac{\partial p}{\partial \lambda} = 0 \qquad (16)$$

where $I_e$: the objective function, information fitness; $d$: the essential cognitive resource; $p$: the physical cost of adding the structural unit; $\eta_c$, $\mu$: Lagrangian factors.

Further, for simplicity it is natural to assume the physical cost of cognitive resources $p(\lambda)$ to be linear with respect to the structural size parameter, measuring the size of the system in terms of effective basic structural cognitive units. Then, the last term in (16) can be assumed to be effectively, a constant, the physical cost of structural variation factor, $\alpha_C$.

From (16), one can estimate the maximum possible gain in the information fitness as:

$$\frac{\partial I_e(\lambda)}{\partial \lambda} \leq \eta_c G_d + \alpha_C \qquad (17)$$

where $G_d = \frac{\partial d}{\partial \lambda}$, the structural gradient of the cognitive resource $d$, $\alpha_C$: the physical cost of structural variation factor.

In one boundary case, the cognitive gradient can be large, for example in cases similar to fully interconnected neural networks, $G_d$ can be proportional to the size of the network: $G_d \propto \lambda$, resulting in an exponential-like boundary on the growth of the information fitness.

In the opposite case, where cognitive resources are severely constrained or the cognitive architecture of the model is ineffective, the cognitive gradient can be diminishing: $G_d \to 0$ and the boundary on the cognitive gain is controlled by the physical factor, with the scenarios of positive linear boundary, $\alpha_C > 0$; stagnation, $\alpha_C \sim 0$; or cognitive decline, $\alpha_C < 0$.

Thus, application of the constrained optimization framework allows to make nontrivial conclusions about development trajectories of emerging intelligent systems with respect to their cognitive architectures.

### 6.3 Collective Intelligence and Information Fitness

The concept of information fitness can be applied to groups or ensembles of natural intelligent systems. For example, it can be instrumental in explaining the effectiveness of the communication strategy widely used by social natural species. Indeed, let us consider a simple hypothetical population of information models *P* of size *N* of a similar architecture with the fitness distribution *p(F)* in the range $R_f = (F_{min}, F_{max})$, $F_{min} \leq F(x) \leq F_{max}$, *x*: an individual model in the population. Suppose $F_a = mean(F(P))$, $F_x = max(F(P))$, the mean and the maximum value of the information fitness in the population.

Then, the total information fitness of the population at some point *t* can then be written as:

$$F_t(P) = \sum_P F(x) = N F_a$$

Next, suppose the population has developed an effective communications strategy that can transfer the information from the best fit information models to the others in the population via transmission of information or "instruction":

$$F(x,t) = C(x, x_{mx}) \sim F_x,$$

where *x*, $x_{mx}$: a common and the best sample in the population; *C*: communication, or instruction function.

Then, the new value of the total population fitness $F_{nt}$ at the next point can be found as:

$$F_{t+1}(P) \sim N F_x(t) = F_t + N (F_x - F_a) = F_t + N G_{com} \qquad (18)$$

where $G_{com} \sim F_x - F_a$: the average individual fitness gain in the population due to communication/instruction.

Then, the improvement in the total fitness of the population due to communication can amplified by up to the size of the population. There are numerous examples of this behavior exhibited by natural biological systems.

An intriguing question then is how such an effective strategy could have emerged in the natural evolution of intelligence. Due to the limitations of the scope and format, this discussion will be deferred to another study.

## 7  Discussion

The connections between the principles of constrained optimization, classical approaches in the theory of natural learning systems and evolutionary processes based on traversal of the extrema landscape of the optimization problem as a strategy for finding the solution, or rather a progression of solutions increasingly approximating the fitness or "adaptation" to the environment examined in this work can be instrumental in a number of ways.

From theoretical perspectives, the results presented here offer a general conceptual and formal mathematical framework for the description of evolving natural intelligent systems. While some interesting initial results have been presented, such as the substantiation of the effectiveness of neural architectures in natural intelligent by the principle of efficient complexity and the classification of incremental adaptations, other qualitative and quantitative results can be expected from developing the proposed formalism.

As mentioned, information-based approaches in the theory of natural intelligence have been developed over a long period since the classical works of Shannon, Shroedinger and others. To discuss or even mention all related studies and directions in one section may not be possible. For this reason, we will limit the discussion to several studies directly related to the subject and scope of this work.

The "information bottleneck" method based on constrained optimization was developed in [19] with the constraint interpreted as the rate distortion function or factor, i.e., in a fundamentally different perspective from the interpretation used in this work. As well, the framework and results of the study apply mostly to the case of learning with classes or categories known at prior (i.e., supervised learning). As was commented, this is rarely the case with natural intelligent systems that cannot rely on prior information and have to "extract" concept structure from their interactions with the environment.

From that perspective, the approaches in the geometric conceptualization of generative representations including those examined in this work, can offer insights into the ability of early intelligent systems to develop effective differentiated behaviors based on grouping similar stimuli into general types or concepts that require a similar response. As noted, the ability to conceptualize sensory stimuli can be critical for natural intelligent systems in meeting the essential physical constraints in the search for optimal architectural solutions.

In [12], the problem of compression of data in unsupervised learning processes and approaches from the information-theoretical and practical perspectives were examined in depth. An observation directly related to the scope of this work is the result on the absence of a natural level or limit of compression in unsupervised learning, in contrast to the supervised case. Then, the practical level of compression is dictated by the essential constraints of natural learning, primarily, those on the memory and computing power that are related directly to the ability of the systems to produce effective differentiated responses via conceptualizing sensory stimuli into manageable frameworks of general types, concepts or external states.

Studies [16,17] among others, used a related approach by defining the objective function in terms of the information entropy. However, they appeared to focus on the maximization of the objective function (entropy), while leaving aside the physical, material constraints of the problem. As we attempted to show in this work, physical constraints, including the critical ones of memory, compute and energy/resources can be of the utmost importance for emerging and developing natural intelligent systems.

Evidently, an unconstrained problem can have essentially different solutions from a strongly constrained one; in fact, even the global minimum of the unconstrained objective function may not satisfy certain practical material constraints. As noted in the introduction section, natural biological systems are by their nature, thermodynamically open, allowing strong flows of both energy and materials; in that setting, the interactions and constraints of energy and materials on the natural systems cannot be ignored. The importance of material, specifically energy constraints of biological intelligent systems is supported by experimental results such as the "critical power law" in neuroscience, the tradeoff between the accuracy of coding and the energy it requires [31].

For these reasons, we believe that the constrained optimization formalism proposed in this work would offer a more accurate description of the states and evolutionary trajectories of progressing natural intelligent systems. Further to its advantage, it is based on the established principles of statistics and information science and does not introduce any new essential assumptions or postulates.

From the more practical point of view, the analysis of necessary conditions of the evolution of natural intelligence, including the principle of "efficient complexity" that is, minimization of the energy cost of architectural adaptations (actually a corollary of the proposed evolutionary formalism) can provide a basis for further quantitative studies into the models and architectures of evolving intelligence.

The concept and framework of analysis based on the information fitness can be instrumental in examination of collective intelligence and intelligent collective behaviors discussed in Section 6; an immediate observation being that behaviors that can effectively propagate or share more effective information models, including communications of information models, can trigger a sharp increase in the information fitness in a collective of learners.

All in all, there seems to be a wide range of directions of research in both theory and applied models of evolving intelligence within the formalism proposed in this work.

## Disclosures

This research has not received any specific funding.

The authors declare no conflicts of interest.